\documentclass[journal]{IEEEtran}

\usepackage{cite}
\usepackage{amsmath,amssymb,amsfonts}
\usepackage{textcomp}
\usepackage{graphicx}
\usepackage[utf8]{inputenc}
\usepackage[T1]{fontenc}
\usepackage{xcolor}
\usepackage{booktabs}
\usepackage{multirow}
\usepackage{array}
\usepackage[font=footnotesize,compatibility=false]{caption}
\usepackage{subcaption}
\usepackage{orcidlink}
\usepackage[ruled,vlined]{algorithm2e}
\SetKw{Return}{return}
\SetAlgoNoLine
\SetAlgoNoEnd
\usepackage{hyperref}
\usepackage{cleveref}

\begin{document}

\title{Dual-Branch INS/GNSS Fusion with Inequality and Equality Constraints}

\author{Mor~Levenhar~\orcidlink{0009-0004-5420-4962},
        Itzik~Klein~\orcidlink{0000-0001-7846-0654}
\thanks{I. Klein and M. Levenhar are with the Autonomous Navigation and Sensor Fusion Lab, Hatter Department of Marine Technologies, University of Haifa, Israel.}}

\maketitle

\begin{abstract}
Reliable vehicle navigation in urban environments remains a challenging problem due to frequent satellite signal blockages caused by tall buildings and complex infrastructure. While fusing inertial reading with satellite positioning in an extended Kalman filter provides short-term navigation continuity, low-cost inertial sensors suffer from rapid error accumulation during prolonged outages. Existing information aiding approaches, such as the non-holonomic constraint, impose rigid equality assumptions on vehicle motion that may be violated under dynamic urban driving conditions, limiting their robustness precisely when aiding is most needed. In this paper, we propose a dual-branch information aiding framework that fuses equality and inequality motion constraints through a variance-weighted scheme, requiring only a software modification to an existing navigation filter with no additional sensors or hardware. The proposed method is evaluated on four publicly available urban datasets featuring various inertial sensors, road conditions, and dynamics, covering a total duration of 4.3 hours of recorded data. Under Full GNSS availability, the method reduces vertical position error by 16.7\% and improves altitude accuracy by 50.1\% over the standard non-holonomic constraint. Under GNSS-denied conditions, vertical drift is reduced by 24.2\% and altitude accuracy improves by 20.2\%. These results demonstrate that replacing hard motion equality assumptions with physically motivated inequality bounds is a practical and cost-free strategy for improving navigation resilience, continuity, and drift robustness without relying on additional sensors, map data, or learned models.
\end{abstract}

\begin{IEEEkeywords}
System state estimation, sensor technology, road transportation, extended Kalman filter, inequality constraints.
\end{IEEEkeywords}

\IEEEpeerreviewmaketitle

\section{Introduction}
Navigation in urban environments has become increasingly important for a wide range of applications, including autonomous driving, public transportation monitoring, and real-time logistics. Reliable vehicle positioning is a fundamental requirement for these systems, as navigation errors directly affect safety, efficiency, and operational performance. At present, global navigation satellite systems (GNSS) are the primary technology used for vehicle positioning due to their global coverage and ability to provide absolute position estimates.\\
However, despite their widespread use, GNSS performance is severely degraded in dense urban areas. Tall buildings, bridges, and other man-made structures frequently obstruct satellite signals, resulting in reduced satellite availability. Under such conditions, GNSS alone cannot maintain reliable positioning, leading to large errors or a complete loss of navigation capability during signal blockages.\\
To address these limitations, GNSS is commonly integrated with an inertial navigation system (INS) using Extended Kalman Filters (EKF) \cite{ibrahim2023enhanced, huang2006low, farrell2008aided, groves2013gnss}. IMUs provide high-rate motion information and are unaffected by external signal obstructions, enabling short-term navigation during GNSS outages. Nevertheless, low-cost IMUs suffer from sensor noise, bias instability, and scale factor errors. These errors accumulate rapidly, leading to significant drift when GNSS updates are unavailable \cite{titterton2004strapdown}. Consequently, maintaining accurate navigation during prolonged GNSS blockages remains a challenging problem.\\
Numerous approaches have been proposed to mitigate these drifts.\\
Many works incorporate additional sensors to constrain vehicle motion and reduce uncertainty, including barometers, magnetometers, and wheel speed sensors \cite{sokolovic2013integration, gao2006development}. Vision-based approaches using cameras \cite{li2022high, gao2022improved, liang2020scalable} and electronic scanning radar \cite{rashed2023enhanced} have also been integrated with inertial-based systems. While effective, these solutions increase system complexity, cost, and computational load. Map-based approaches further introduce constraints using digital road networks or geographic databases \cite{betaille2010creating, levinson2010robust}, but they rely heavily on accurate and up-to-date map information, which is not always available.\\
An alternative to adding additional sensors and specific hardware is information aiding. Information aiding exploits prior knowledge of the platform's motion characteristics to generate pseudo-measurements that update the navigation filter without requiring additional physical sensors ~\cite{engelsman2023information}. Unlike external sensor aiding, it is continuously available at the INS sampling rate at no hardware cost, which makes it particularly well-suited for urban navigation scenarios where GNSS outages are frequent, and the deployment of additional sensors is impractical \cite{ma2003vehicle}. These include estimating lateral velocity \cite{gao2022improved}, enforcing equality constraints such as zero lateral velocity or constant height assumptions \cite{klein2011vehicle}, or maintaining constant slope through pseudo-measurements \cite{klein2010pseudo}. Additional information can also be extracted by incorporating the known geometric relationship between sensor mounting points as a virtual lever-arm measurement \cite{borko2018gnss}. Furthermore, previously accumulated position updates can be re-exploited within the navigation filter to enhance heading estimation, effectively squeezing additional information from measurements that would otherwise be used only once \cite{klein2020squeezing}.\\
Some studies focus on enhancing motion models by exploiting road geometry, such as road curvature \cite{tzoreff2011novel}, or by representing vehicle trajectories as sequences of straight line segments \cite{cui2003autonomous}. While these methods introduce additional structure, their performance is often limited by modeling assumptions that may not hold in complex, unstructured urban environments.\\
Other studies simplify the state representation based on land vehicle dynamics \cite{toledo2006imm, yang2017comparison}. More recently, hybrid frameworks combining EKF-based sensor fusion with deep learning have been proposed \cite{yao2017hybrid, zhang2019fusion, liu2021vehicle}, though these often require extensive training data and may lack interpretability.\\
In this paper, we propose a novel information-aided INS/GNSS integrated navigation framework designed to mitigate the limitations of standard kinematic constraints in urban environments. The core of the proposed method lies in a dual-branch estimation architecture: one branch applies standard non-holonomic constraints (NHC) using equality assumptions, while the second branch rigorously enforces inequality constraints \cite{gupta2007kalman} derived from the physical limitations of the vehicle, such as maximum velocity, roll, pitch, and height variations. This approach allows the filter to accommodate natural vehicle motion while effectively preventing the solution from drifting into physically impossible states. The final state estimate is obtained through a variance-weighted fusion of these two branches, ensuring that the navigation solution remains both dynamic and bounded.\\
The main contributions of this work are threefold:
\begin{enumerate}
    \item The development of an information-aided methodology for embedding inequality constraints into the INS/GNSS estimation framework. Unlike equality-based pseudo-measurements, the proposed formulation imposes no assumption on the nominal motion state, making it applicable across diverse driving conditions without model-specific tuning.
    \item A dual-branch pipeline that integrates both equality and inequality constraints, enabling the simultaneous handling of multiple constraint types within a unified framework. 
    \item Demonstration of the robustness of the proposed method across four diverse datasets recorded by different platforms under both full GNSS availability and GNSS-denied conditions.
\end{enumerate}
Experimental validation on four publicly available urban datasets, with a total duration of 4.3h, demonstrates that the proposed method consistently outperforms other approaches under full GNSS availability and GNSS-denied conditions. Notably, under GNSS-denied conditions, the proposed method reduces the 95th percentile vertical position error by 42.5\%, demonstrating strong robustness against extreme altitude drift events.\\
The remainder of this paper is organized as follows. Section II presents the system model and problem formulation. Section III provides a detailed description of the proposed EKF framework, incorporating both equality and inequality constraints to bound navigation drift. Section IV discusses the experimental setup and provides a comprehensive analysis of the results across various urban datasets. Finally, concluding remarks are provided in Section V.\\
For convenience, all abbreviations used throughout this paper are listed in Table~\ref{tab:abbreviations}.
\begin{table}[tb]
\centering
\caption{Abbreviations and Descriptions.}
\label{tab:abbreviations}
\renewcommand{\arraystretch}{1.2}
\begin{tabular}{ll}
\toprule
\textbf{Abbreviation} & \textbf{Description} \\
\midrule
ARMSE & Altitude RMSE \\
ECEF  & Earth-Centered Earth-Fixed \\
EKF   & Extended Kalman Filter \\
es-EKF & Error-State Extended Kalman Filter \\
GNSS  & Global Navigation Satellite System \\
GPS   & Global Positioning System \\
GT    & Ground Truth \\
IMU   & Inertial Measurement Unit \\
INS   & Inertial Navigation System \\
NED   & North-East-Down \\
NHC   & Non-Holonomic Constraint \\
PRMSE & Position RMSE \\
QP    & Quadratic Programming \\
RMSE  & Root Mean Square Error \\
RTK   & Real-Time Kinematic \\
VRMSE & Velocity RMSE \\
\bottomrule
\end{tabular}
\end{table}
\section{Problem Formulation}
The nonlinear nature of the INS equations of motion requires a 
nonlinear filter. The most widely used approach for fusing INS 
With external aiding sensors, the error-state extended Kalman 
filter (es-EKF) \cite{farrell2008aided}. For a low-cost IMU, a 15-element 
error-state vector is defined as:
\begin{equation}
    \boldsymbol{\delta} \mathbf{x} = \begin{bmatrix}
        \boldsymbol{\delta} \mathbf{p}^{n} & \boldsymbol{\delta}^{n} \mathbf{v} & \boldsymbol{\epsilon}^{n} & \mathbf{\delta b_a}^{b} & \mathbf{\delta b_g}^{b}
    \end{bmatrix} ^{\mathsf{T}} \in \mathbf{R}^{15}
    \label{eq:error_state_vector}
\end{equation}
where, $\boldsymbol{\delta} \mathbf{p}^{n}$ is the position error state vector expressed in navigation frame, $\boldsymbol{\delta}^{n} \mathbf{v}$ is the velocity error state expressed in navigation frame, $\boldsymbol{\epsilon}^{n}$ is the altitude error state expressed in navigation frame, $\mathbf{\delta b_a}^{b}$ is the accelerometer bias error vector expressed in body frame and $\mathbf{\delta b_g}^{b}$ is the gyro bias error vector expressed in body frame.\\
It is worth noting that, among the three orientation angles, yaw ($\psi$) is weakly observable in land vehicle navigation when GNSS provides only position updates, as no direct measurement of heading is available from the aiding source \cite{hong2005observability}.
\subsection{Kalman Prediction Phase}
The error state vector is defined as:
\begin{equation}
    \mathbf{\delta x} = \tilde{\mathbf{x}} - \mathbf{x}
    \label{eq:dx}
\end{equation}
where, $\mathbf{\delta x}$ is the error state vector, $\tilde{\mathbf{x}}$ is the estimate state vector and $\mathbf{x}$ is the true nominal state.\\ 
The linearized inertial error model is defined by \cite{farrell2008aided}:
\begin{equation}
    \dot{\mathbf{\delta x}} = \mathbf{F}\cdot \boldsymbol{\delta}\mathbf{x}+\mathbf{G}\cdot \boldsymbol{\omega}
    \label{eq:general_model_error}
\end{equation}
where, $\boldsymbol{\omega} = \begin{bmatrix} \omega_{ax}^{\mathsf{T}} & \omega_{ay}^{\mathsf{T}} & \omega_{az}^{\mathsf{T}} & \omega_{gx}^{\mathsf{T}} & \omega_{gy}^{\mathsf{T}} & \omega_{gz}^{\mathsf{T}}\end{bmatrix}^{\mathsf{T}}$ is the accelerometer and gyro noise, and the full definitions of $\mathbf{F}$ and $\mathbf{G}$ can be found in standard navigation textbooks \cite{groves2013gnss, farrell2008aided, titterton2004strapdown}.\\
The covariance matrix is propagated as:
\begin{equation}
    \mathbf{P}_{k+1}^- = \mathbf{\Phi}_k \mathbf{P}_k^+ \mathbf{\Phi}_k^\top + \mathbf{\Gamma} \mathbf{Q} \mathbf{\Gamma} ^{\mathsf{T}}
    \label{eq:prior_covariance}
\end{equation}
where $\textbf{Q}$ is the process-noise covariance matrix, $\mathbf{\Phi} = e^{\mathbf{F}(t)\Delta t}$ and $\mathbf{\Gamma} \mathbf{Q}_k \mathbf{\Gamma} ^{\mathsf{T}} \approx \mathbf{G}\mathbf{Q}_k\mathbf{G}^{\mathsf{T}}\Delta t$, as the inertial sensor operates in high sampling  rate. The superscripts $^-$ and $^+$ represent estimates before and after incorporating measurements.
\subsection{Kalman Update Phase}
The measurement model is defined as:
\begin{equation}
    \mathbf{y}_k = \mathbf{H} \mathbf{x}_{k} + \boldsymbol{\nu}_{k}
    \label{eq:yk}
\end{equation}
and its error model for the es-EKF is:
\begin{equation}
    \delta \mathbf{y}_k = \mathbf{H} \, \delta \mathbf{x}_{k} + \boldsymbol{\nu}_{k}
    \label{eq:delta_yk}
\end{equation}
where $\mathbf{H}$ is the measurement matrix that maps the state vector to the measurement space, $\mathbf{y}_k$ represents the measurement value at time $k$, $\delta \mathbf{y}_k$ and $\boldsymbol{\nu}_k$ represent the output error and the measurement noise, respectively. The covariance matrix of $\boldsymbol{\nu}_k$ is denoted as $\mathbf{R}$.\\
When considering position updates from GNSS, the measurement matrix $\mathbf{H} \in \mathbb{R}^{3 \times 15}$ is:
\begin{equation}
    \mathbf{H} = \begin{bmatrix}
    \mathbf{I}_{3 \times 3} & 0_{3 \times 3} & 0_{3 \times 3} & 0_{3 \times 3}& 0_{3 \times 3}
    \end{bmatrix}
    \label{eq:H_exmple}
\end{equation}
and $\mathbf{\delta y}_k$ is a vector of the position error calculated from the GNSS.\\
\begin{equation}
    \mathbf{\delta y} = \begin{bmatrix}
        \hat{N} - N_{\text{GNSS}} \\
        \hat{E} - E_{\text{GNSS}} \\
        \hat{h} - h_{\text{GNSS}}
    \end{bmatrix}
\end{equation}
where $\hat{N}$, $\hat{E}$, and $\hat{h}$ denote the estimated North, East, and height coordinates, while $N_{\text{GNSS}}$, $E_{\text{GNSS}}$, and $h_{\text{GNSS}}$ represent the corresponding measurements from the low-cost GNSS receiver.\\
The Kalman gain is:
\begin{equation}
    \mathbf{K}_k = \mathbf{P}_k^- \mathbf{H}_k^\top \left( \mathbf{H}_k \mathbf{P}_k^- \mathbf{H}_k^\top + \mathbf{R} \right)^{-1}
    \label{eq:kalman-gain}
\end{equation}
Finally, the updated error state is:
\begin{equation}
    \mathbf{\delta\hat{x}}_k^+ = \mathbf{K}_k \cdot \mathbf{\delta y}_k
    \label{eq:Xk+1+}
\end{equation}
and the covariance matrix is updated using: 
\begin{equation}
\begin{aligned}
\hat{\mathbf{P}}^{+}_{k}
&=
(\mathbf{I}-\mathbf{K}_k\mathbf{H}_k)\mathbf{P}_{k}^{-}
(\mathbf{I}-\mathbf{K}_k\mathbf{H}_k)^{\mathsf{T}}
\\
&\quad
+ \mathbf{K}_k\mathbf{R}\mathbf{K}_k^{\mathsf{T}}
\end{aligned}
\label{eq:P+}
\end{equation}

\subsection{Error-State Correction}

After the Kalman filter measurement update, the estimated error-state vector
\(
\delta \hat{\mathbf{x}}_k^+
\)
is used to correct the nominal navigation state, using the following equations \cite{groves2013gnss}:
\begin{align}
\mathbf{p}_{k}^{+} &= \mathbf{p}_{k}^{-} - \delta \mathbf{p}_k^+ \label{eq:update_p}\\
\mathbf{v}_{k}^{+} &= \mathbf{v}_{k}^{-} - \delta \mathbf{v}_k^+ \label{eq:update_v}\\
\mathbf{T}_{k}^{+} &=
(\mathbf{I}_3
+
[ \boldsymbol{\epsilon}_k^+]_\times)^\mathsf{T} \, \mathbf{T}_{k}^{-}
\label{eq:update_t}\\
\mathbf{b}_{k}^{+} &=
\mathbf{b}_{k}^{-}
+
\delta \mathbf{b}_k^+
\label{eq:update_b}
\end{align}
\subsection{Information-Aided Constraints}
The NHC represents a class of motion constraints that restrict the motion of wheeled vehicles due to rolling-without-slipping conditions. In navigation systems, NHCs are commonly used as information-aided constraints to improve state observability during GNSS outages.\\
A widely adopted realization of the NHC in inertial navigation assumes that, during nominal ground-vehicle motion, the lateral and vertical velocity components in the body frame are zero. These constraints are typically incorporated as pseudo-measurements within the state estimation framework
\cite{engelsman2023information}. To apply this constraint, we define the measurement matrix $\mathbf{H}_{\mathrm{NHC}} \in \mathbb{R}^{2 \times 15}$ as
\begin{multline}
\mathbf{H}_{\mathrm{NHC}}
=
\begin{bmatrix}
\mathbf{e}_2^{\mathsf{T}} \\
\mathbf{e}_3^{\mathsf{T}}
\end{bmatrix}
\begin{bmatrix}
\mathbf{0} &
\mathbf{T}_n^b &
-\mathbf{T}_n^b \,[\mathbf{v}^n]_\times &
\mathbf{0} &
\mathbf{0}
\end{bmatrix}
\end{multline}
where $\mathbf{e}_2$ and $\mathbf{e}_3$ denote the second and third standard basis vectors of $\mathbb{R}^3$, respectively.

\section{Proposed Methodology}
\begin{figure*}
    \centering
    {\includegraphics[width=0.99\linewidth]{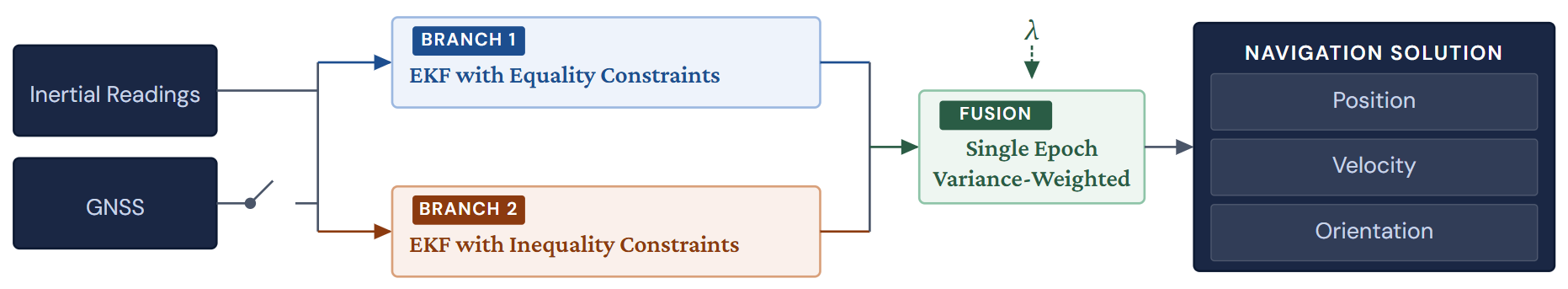}}
    \caption{Proposed dual-branch EKF architecture. Inertial readings are always available and feed both branches continuously. GNSS input is switchable, allowing both branches to operate with or without satellite positioning.}
    \label{fig:proposed methodology diagram}
\end{figure*}

To improve INS/GNSS fusion in normal operating conditions and in GNSS-denied scenarios, only using information aiding, we propose a dual-branch fusion architecture that combines equality and inequality constraints derived from vehicle body dynamics. As illustrated in Fig.~\ref{fig:proposed methodology diagram}, the framework maintains two parallel estimation branches, each exploiting a fundamentally different type of physical knowledge.
The first branch applies the non-holonomic constraint (NHC) as an equality pseudo-measurement, encoding the kinematic assumption that a ground vehicle exhibits zero lateral and vertical body-frame velocity during nominal motion \cite{engelsman2023information}. While effective under standard driving conditions, this assumption is rigid. Thus, any violation, such as skidding or abrupt maneuvers, directly degrades the filter solution. The second branch enforces inequality constraints derived from the physical operating envelope of the vehicle, bounding height, attitude, and forward velocity within known feasible limits. Unlike equality constraints, these bounds impose no assumption about the nominal state; they only prevent the solution from entering physically impossible regions.
Applying both constraint types within a single filter would raise a conflict when the equality assumption is momentarily violated. Instead, we maintained the two branches independently, and their estimates are fused through a variance-weighted scheme, allowing each to contribute according to its instantaneous reliability. This architecture preserves the interpretability and stability of the model-based EKF while ensuring that the navigation solution remains physically bounded throughout the trajectory.\\
In the following subsection, we elaborate on each of the two branches and the fusion process.\\
\subsection{Motion-Derived Inequality Constraints}
The inequality constraints are derived and implemented differently in two scenarios: 1) only the inequality branch operation, and 2) in the dual-branch fusion.
A quadratic programming approach is employed in the single-branch operation when only the inequality constraints apply, mainly in GNSS-denied scenarios or between two successive GNSS updates. In the dual-branch scenarios, where a combination of equality constraints, GNSS, and inequality constraints exists, a Kalman gain optimization approach \cite{gupta2007kalman} is applied.
For both scenarios, the inequality constraints are expressed in the form of an equation \eqref{eq:inequla_constraint2}.\\
\subsubsection{\textbf{Inequality Constraint Only}} When only inequality constraints exist, we first calculate the unconstrained state estimate and then project it onto the feasible region defined by the inequality constraints. This projection is formulated as a quadratic programming (QP) problem that minimizes the distance between the unconstrained and constrained estimates while ensuring the inequality constraints are satisfied \cite{gupta2007kalman}:
\begin{equation}
\mathbf{\hat{x}}_k^+ = \underset{\mathbf{\hat{x}}_k^+}{\mathrm{argmin}}~(\mathbf{\hat{x}}_k^+-\mathbf{\hat{x}}_k^-)^{\mathsf{T}} \tilde{\mathbf{P}}^{-1} (\mathbf{\hat{x}}_k^+-\mathbf{\hat{x}}_k^-)
\label{eq:x+_ineqlity}
\end{equation}
subject to
\begin{equation}
\mathbf{C}\mathbf{\hat{x}}_k^+ \leq \mathbf{d}
\label{eq:inequla_constraint2}
\end{equation}
where $\mathbf{C} \in \mathbb{R}^{l \times n}$ defines the linear relationships between state variables that form the inequality constraints, $\mathbf{d} \in \mathbb{R}^{l}$ is a vector of constraint limits, and $l$ and $n$ represent the number of inequalities constraints and the number of state variables, respectively.\\
\subsubsection{\textbf{inequality constraint with additional measurements}} When additional measurements (from GNSS) together with inequality constraints (from vehicle motion) exist, we follow a two-step process. Initially, we calculate the standard Kalman gain using only the GNSS measurements, as detailed in~\eqref{eq:kalman-gain}. In the second step, we refine the Kalman gain to ensure the state estimate also satisfies the inequality constraints. This involves solving an optimization problem that minimizes the estimation covariance while ensuring both equality and inequality constraints are satisfied. After the gain optimization, we follow the standard EKF procedure to update the error state and its covariance matrix.\\
The Kalman gain optimization problem is defined as \cite{gupta2007kalman}:
\begin{gather}
\mathbf{K}_k
=
\underset{\mathbf{K}_d \in \mathbb{R}^{n\times m}}{\mathrm{argmin}}\;
\mathrm{trace}\Big[ \Delta \Big]\\
\Delta = (\mathbf{I}-\mathbf{K}_d\mathbf{H}_k)\mathbf{P}_{k}^-(\mathbf{I}-\mathbf{K}_d\mathbf{H}_k)^{\mathsf{T}}
+\mathbf{K}_d\mathbf{R}\mathbf{K}_d^{\mathsf{T}}
\label{eq:optimal_K_inequal}
\end{gather}
subject to
\begin{equation}
\begin{aligned}
\mathbf{H}(\mathbf{\hat{x}}^+_{(\mathbf{K_d})}) = \mathbf{y} \\
\mathbf{C}(\mathbf{\hat{x}}^+_{(\mathbf{K_d})}) \leq \mathbf{d}
\end{aligned}
\label{eq:inequality}
\end{equation}
where $\mathbf{K}_d$ is the initial value of $\mathbf{K}_k$, considering only the equality constraint, and $\mathbf{\hat{x}}^+_{(\mathbf{K_d})}$ is estimated following \eqref{eq:Xk+1+}, \eqref{eq:update_p}-\eqref{eq:update_b}.\\
The detailed implementation of the Kalman gain optimization is provided in algorithm~\ref{alg:kalman_simulation}.\\
\subsubsection{\textbf{Inequality Branch Implementation}} Four inequality constraints are proposed to bound the physically admissible region of the state estimate: height, roll ($\phi$), pitch ($\theta$), and forward velocity.\\
For the height, roll, and pitch, the inequality matrix $C \in \mathbb{R}^{6 \times 15}$ and the limits vector $\mathbf{d} \in \mathbb{R}^{6}$ are constant over time and have the same format in both the QP approach and the Kalman gain optimization:
\begin{equation}
    \mathbf{C} = 
    \begin{bmatrix}
        \mathbf{0}_{1 \times 2} & 1 & \mathbf{0}_{1 \times 12} \\
        \mathbf{0}_{1 \times 6} & 1 & \mathbf{0}_{1 \times 8} \\
        \mathbf{0}_{1 \times 7} & 1 & \mathbf{0}_{1 \times 7} \\
        \mathbf{0}_{1 \times 2} & -1 & \mathbf{0}_{1 \times 12} \\
        \mathbf{0}_{1 \times 6} & -1 & \mathbf{0}_{1 \times 8} \\
        \mathbf{0}_{1 \times 7} & -1 & \mathbf{0}_{1 \times 7} \\
    \end{bmatrix}
    \label{eq:c_hpr}
\end{equation}
and
\begin{equation}
\mathbf{d} =
\begin{bmatrix}
h_{\max} \\
\phi_{\max}\\
\theta_{\max} \\
- h_{\min} \\
- \phi_{\min} \\
- \theta_{\min}
\end{bmatrix}
\label{eq:d_hpr}
\end{equation}
The forward velocity constraint takes different forms in the two operational scenarios due to a fundamental difference in their optimization domains. In the Kalman gain optimization case, the constraint is imposed through the measurement matrix structure and is therefore naturally expressed in the body frame using the transformation matrix $\mathbf{T}^b_n$, as follows:
\begin{multline}
\mathbf{C}
=
\begin{bmatrix}
\mathbf{e}_1^{\mathsf{T}} \\
- \mathbf{e}_1^{\mathsf{T}}
\end{bmatrix}
\begin{bmatrix}
\mathbf{0} &
\mathbf{T}_n^b &
-\mathbf{T}_n^b \,[\mathbf{v}^n]_\times &
\mathbf{0} &
\mathbf{0}
\end{bmatrix}
\label{C_non_linear}
\end{multline}
where $\mathbf{e}_1$ denotes the first standard basis vectors of $\mathbb{R}^3$
\begin{equation}
    d = \begin{bmatrix}
        v_{max}\\
        v_{max}
    \end{bmatrix}
    \label{eq:d_v_max}
\end{equation}
and $v_{\max}$ is a hyperparameter bounding the forward velocity, such that $|v_k|\le v_{\max}$ at all times. This formulation enforces the inequality constraint directly on the forward velocity in the body frame.\\
In the QP projection case, however, the optimization operates directly on the navigation-frame state vector. Expressing the body-frame forward velocity in navigation-frame coordinates would require embedding the rotation matrix $\mathbf{T}^b_n$ into the constraint matrix $\mathbf{C}$, rendering the constraint nonlinear and incompatible with the linear structure required by~\eqref{eq:x+_ineqlity}--\eqref{eq:inequla_constraint2}. 
Instead, we bound the vertical navigation-frame velocity component $v_d$, which is geometrically related to the forward velocity through the pitch angle:
\begin{equation}
    v_{d,\max} = \left| \sin(\theta_k) \right| \cdot v_{\max}
    \label{eq:vd_max}
\end{equation}
where $\theta_k$ denotes the pitch angle at epoch $k$. This formulation preserves the linearity of the constraint while enforcing an equivalent physical bound. The corresponding constraint matrix and the bound vector are then:
\begin{equation}
    \mathbf{C} = \begin{bmatrix}
        \mathbf{0}_{1 \times5} & 1 & \mathbf{0}_{1 \times9} \\
        \mathbf{0}_{1 \times5} & -1 & \mathbf{0}_{1 \times9}
    \end{bmatrix}
    \label{eq:C_v}
\end{equation}
\begin{equation}
    \mathbf{d} = \begin{bmatrix}
        v_{d,\max} \\
        v_{d,\max}
    \end{bmatrix}
    \label{eq:d_v}
\end{equation}
\begin{algorithm}
\caption{Kalman Filter with Inequality Constraints with and without GNSS (Branch 2, Figure \ref{fig:proposed methodology diagram}).}
\label{alg:kalman_simulation}
\SetKwInOut{Input}{Input}
\SetKwInOut{Output}{Output}
\SetKw{KwStep}{Step:}
\Input{IMU measurements $(\mathbf{f},\boldsymbol{\omega}) \in \mathbb{R}^{N}$\\
    initial state $\hat{\mathbf{x}}_0 \in \mathbb{R}^{n}$\\
    initial error-state $\delta\hat{\mathbf{x}}_0 \in \mathbb{R}^{n}$\\
    initial covariance $\mathbf{P}_0 \in \mathbb{R}^{n\times n}$\\
    process-noise covariance $\mathbf{Q} \in \mathbb{R}^{n\times n}$\\
    measurement $\mathbf{y} \in \mathbb{R}^{m}$\\
    measurement matrix $\mathbf{H} \in \mathbb{R}^{m\times n}$\\
    measurement noise covariance $\mathbf{R} \in \mathbb{R}^{m \times m}$\\
    inequality constraints $(\mathbf{C},\mathbf{d})$%
}
\Output{$\{\hat{\mathbf{x}}\},\{\hat{\mathbf{P}}\}$}
\For{$k=1$ \KwTo $N-1$}{
    Propagate $\mathbf{x}_k$ using $\mathbf{f}_k,\boldsymbol{\omega}_k$ and $\Delta t$ \;
    $\mathbf{P}_k^- \leftarrow \boldsymbol{\Phi}\mathbf{P}_{k-1}\boldsymbol{\Phi}^\mathsf{T}+\mathbf{Q}$\;
    \uIf{valid GNSS}{
        $\delta\hat{\mathbf{x}}^- \leftarrow \mathbf{0}$\;
        $\mathbf{K} \leftarrow \mathbf{P}_k^- \mathbf{H}_k^\top \left( \mathbf{H}_k \mathbf{P}_k^- \mathbf{H}_k^\top + \mathbf{R}_k \right)^{-1}$\;
        $\delta\hat{\mathbf{x}}^+ \leftarrow \delta\hat{\mathbf{x}}^- + \mathbf{K}\,\boldsymbol{\delta y}$\;
        $\mathbf{x}^+_k \leftarrow \mathrm{StateCorrection}(\mathbf{x}^+_{k-1},\delta\hat{\mathbf{x}}^+,\mathbf{T}_k)$\;
        \If{$\exists\, i:\; \mathbf{C}\mathbf{x}^+_{k,i} > \mathbf{d}_i + \epsilon$}{
            $n_c \leftarrow \mathrm{length}(\mathbf{d})$\;
            $\mathcal{G} \leftarrow \emptyset$\;
            \For{$i = 0$ \KwTo $n_c-1$}{
                $\mathbf{c}^* \leftarrow \mathbf{C}_{i,:}$\;
                $\mathcal{G} \leftarrow \mathcal{G} \cup \bigl\{g(\mathbf{K})= \mathrm{Constraint}(
                \mathbf{K}, \mathbf{c}^*, d_i, \mathbf{y}, \mathbf{x}, \mathbf{T}
                ) \bigr\}$ \eqref{eq:inequality}\;
            }
            $\mathbf{K}_c \leftarrow \arg\min\limits_{\mathbf{K}}\;\mathrm{Objective}(\mathbf{K},\mathbf{H},\mathbf{P},\mathbf{R})$
            $\text{ s.t. } g(\mathbf{K}) \ge 0,\ \forall g \in \mathcal{G}$ \eqref{eq:optimal_K_inequal}\;
            \If{solver failed}{
                $\mathbf{K}_c \leftarrow \mathbf{K}$\;
            }
        }
        \Else{
            State already satisfies constraints\;
        }
        $\delta\hat{\mathbf{x}}^{+} \leftarrow \mathbf{K}_c\,\boldsymbol{\delta y}$\;
        $\mathbf{P}^{+}_k \leftarrow (\mathbf{I}-\mathbf{K}_c\mathbf{H}_k)\mathbf{P}_{k}^-(\mathbf{I}-\mathbf{K}_c\mathbf{H}_k)^{\mathsf{T}}+\mathbf{K}_c\mathbf{R}\mathbf{K}_c^{\mathsf{T}}$\;
        $\hat{\mathbf{x}}_k \leftarrow \mathrm{StateCorrection}(\mathbf{x}^+_{k-1},\delta\hat{\mathbf{x}}^+,\mathbf{T}_k)$\;
        $\hat{\mathbf{P}}_k \leftarrow \mathbf{P}^{+}_k$\;
    }
    \Else{
        $\hat{\mathbf{x}}_k \leftarrow \texttt{optimizer}(\mathbf{x}_k,\mathbf{P}_k^{-},\mathbf{C},\mathbf{d})$
        solving \eqref{eq:x+_ineqlity}--\eqref{eq:inequla_constraint2}\;
        $\hat{\mathbf{P}}_k \leftarrow \mathbf{P}^{-}_k$\;
    }
    Store $\hat{\mathbf{x}}_k,\hat{\mathbf{P}}_k$\;
}
\Return{$\{\hat{\mathbf{x}}\},\{\hat{\mathbf{P}}\}$}\;
\end{algorithm}
\subsection{Dual-Branch Fusion}
To integrate the results of the two branches, we employ a single-epoch fusion algorithm \cite{groves2013gnss}. In this approach, a variance-weighted average is computed between the two branches' solutions. The weights are determined by the estimated variances and are computed separately for each state variable at every epoch.\\
To bias the fusion process toward the inequality-constrained solution, we introduce a constant vector $\boldsymbol{\lambda}$ that amplifies the weights of the bounded states (height, roll, pitch, and vertical velocity) and reduces the weights of the horizontal position components.\\
The values of $\boldsymbol{\lambda}$ are motivated by a fundamental asymmetry in the covariance information available from each branch. In the equality-constrained branch, the Kalman update step produces a posterior covariance $\mathbf{P}_{NHC}^+$ that reflects the full measurement correction, providing a reliable uncertainty estimate for all state variables. In the inequality-constrained branch, however, the QP projection does not perform a standard Kalman update as no measurement equation is linearized and no Kalman gain is computed. As a result, the covariance matrix $\mathbf{P}_{INQ}^+$ is not updated during the projection step and retains the prior covariance $\mathbf{P}_k^-$, causing the uncertainty of the bounded states to be overestimated relative to their true post-correction values. Consequently, a naive variance-weighted fusion would systematically underweight the inequality-constrained solution for precisely those state variables that the projection most effectively constrains. To compensate for this covariance inconsistency, the elements of $\boldsymbol{\lambda}$ corresponding to the bounded states are set to values greater than unity, amplifying their contribution in the fusion step. The remaining elements are set to values below unity to reduce the influence of the inequality branch on unconstrained states, where the NHC solution provides a more reliable estimate. This design ensures that the fusion weights reflect the true relative reliability of each branch, compensating for the absence of a covariance update in the QP projection step.\\
We calculate the weights using:
\begin{equation}
    w_{j}^{INQ} 
    = \left( 
    \sum_{i=1}^{2} 
    \mathbf (1/(\sigma_{j}^i)^2)
    \right)\cdot (\sigma^{INQ}_{j})^2\cdot \lambda_{j}
    \label{eq:w_propoused_inq}
\end{equation}
\begin{equation}
    w_j^{NHC}=1- w_j^{INQ}
    \label{eq:w_propoused_nhc}
\end{equation}
where $i$ indexes the branch solution, $i \in \{1,2\}$ corresponding to the inequality-constrained/GNSS (branch 2) and NHC/GNSS (branch 1),
$j$ indexes the state variable, $j \in \{1,2,3,\dots,15\}$, 
and $\sigma_{ij}$ denotes the estimated standard deviation associated with the $j$-th state variable in branch $i$.\\
Finally, the estimated state is:
\begin{equation}
    {\mathbf{\hat{x}}_k} 
    = \sum_{i=1}^2 (\mathbf{w}^i \cdot \mathbf{x}_k^{i+})
    \label{eq:x_propoused}
\end{equation}
where $\mathbf{\hat{x}_k}$ is the estimated state at time $k$. 
\section{Analysis and Results}

\subsection{Datasets}
We used four publicly available datasets that include inertial measurements (specific force and angular velocity), with ground truth (GT) obtained using RTK-GNSS:
\begin{enumerate}
    \item \textbf{KITTI}: A large-scale autonomous driving dataset captured from a VW Passat station wagon equipped with stereo cameras, a Velodyne 3D laser scanner, and a GPS/IMU system, recorded across diverse real-world traffic scenarios in and around Karlsruhe, Germany \cite{Geiger2013IJRR}. We used all sequences with a duration longer than 1.5 minutes, resulting in 12 sequences with a total duration of approximately 55 minutes. The measurements were collected using an OXTS RT3003 inertial and GNSS navigation system (position accuracy: 0.02 m, altitude accuracy: 0.1°).
    \item \textbf{TOKYO}: Collected using an instrumented vehicle platform across two urban environments: Odaiba and Shinjuku \cite{hsu2021urbannav}. We used both sequences of this dataset, with a total duration of approximately one hour. Ground truth was collected using an Applanix POS LV 620 system, while the inertial data were recorded using a Tamagawa TAG264 IMU.
    \item \textbf{HONGKONG}: A multisensory collection gathered using a ground vehicle platform (Honda Jazz/Fit) equipped with a roof-mounted sensor kit, capturing navigation data across diverse urban environments in Hong Kong \cite{hsu2023hong}. We used three sequences (excluding the tunnel sequence), with a total recording time of approximately 1.3 hours. Ground truth was recorded using a NovAtel SPAN-CPT system processed with Inertial Explorer, while the inertial data were recorded using an Xsens IMU.
    \item \textbf{ODYSSEY}: An automotive lidar-inertial odometry (LIO) dataset recorded from a car-mounted platform \cite{kurda2025odyssey}. We used four different routes, each repeated three times, with a total recording time of approximately 1.5 hours. The trajectories were named \emph{Bertway}, \emph{CountryRoad}, \emph{InnerCity}, and \emph{Theater}. Position GT was recorded using the iPRENA-M-II, an RLG-based navigation-grade INS. Velocity ground truth was obtained by differentiating the position measurements with respect to time in the NED coordinate frame. The inertial data were recorded using the iNAT M300-TLE-LN1, an automotive-grade MEMS-based INS.
\end{enumerate}
In total, we evaluated the proposed algorithm on four datasets, comprising 4.3 hours of data collected across 29 sequences and covering a total distance of 93 km, with varying sampling frequencies. A summary of the dataset characteristics is provided in Table~\ref{tab:dataset_parameters}.

\begin{table*}[htbp]
\centering
\caption{Main Dataset Parameters.}
\label{tab:dataset_parameters}
\begin{tabular}{lccccc}
\toprule
\textbf{DATASET} & \textbf{Number of Trajectories} & \textbf{TOTAL DISTANCE (km)} & \textbf{TOTAL TIME (min)} & \textbf{AVG. HEIGHT RANGE (m)} & \textbf{IMU Rate (Hz)} \\
\midrule
Hongkong & 3 & 11.20 & 77.4 & 8.73 & 100 \\
Odyssey & 12 & 43.05 & 82.2 & 29.96 & 250 \\
Tokyo & 2 & 14.28 & 55.6 & 11.59 & 50 \\
Kitti & 12 & 24.69 & 45.4 & 20.63 & 100 \\
\bottomrule
\end{tabular}
\end{table*}
\subsection{Evaluation Metrics}
To evaluate the performance of the proposed method, we use the root mean square error (RMSE) and the 95th percentile error.\\
We calculate the RMSE for the position, velocity and altitude. The position RMSE (PRMSE) is defined by:
\begin{equation}
\mathrm{PRMSE} =
\sqrt{\frac{1}{N} \sum_{k=1}^{N}
\left\| \hat{\mathbf{p}}_k - \mathbf{p}_k^{\mathrm{GT}} \right\|^2}
\label{eq:prmse}
\end{equation}
where $N$ is the total number of epochs , $\hat{\textbf{p}}_{k}$ is the estimated position vector at epoch k, and $\textbf{p}_k^{\mathrm{GT}}$ denotes the position GT value at epoch $k$.
The velocity RMSE (VRMSE) is defined by:
\begin{equation}
\mathrm{VRMSE} =
\sqrt{\frac{1}{N} \sum_{k=1}^{N}
\left\| \mathbf{\hat{v}}_k - \mathbf{v}_k^{\mathrm{GT}} \right\|^2 }
\label{eq:vrmse}
\end{equation}
where $\mathbf{\hat{v}}_k$ is the estimated velocity vector at epoch k, $\mathbf{v}_k^{\mathrm{GT}}$ represents the velocity GT at epoch k .
Lastly, the altitude RMSE (ARMSE) is defined by $\phi$ and $\theta$:
\begin{equation}
\mathrm{ARMSE} =
\sqrt{\frac{1}{N} \sum_{k=1}^{N}
\left\| \begin{bmatrix}
    \hat{\phi} \\
    \hat{\theta}
\end{bmatrix}_k - \begin{bmatrix}
    \phi \\
    \theta
\end{bmatrix}_k^{\mathrm{GT}} \right\|^2 }
\label{eq:armse}
\end{equation}
Where, $\phi$ and $\theta$ represent roll and pitch angles, respectively.
Yaw is excluded from this evaluation as discussed in Section~II.\\
In addition to RMSE, we report the 95th percentile error, defined as the value below which 95\% of the absolute errors fall. This metric highlights robustness and rare large deviations, particularly under GNSS outages.\\
\subsection{Comparison with Existing Methods}
The proposed dual-branch fusion is compared against three approaches:
\begin{enumerate}
    \item \textbf{EKF}: This is the baseline approach, without motion constraints, relying solely on GNSS position updates.
    \item \textbf{NHCEKF}: Same as the EKF approach but with augmented non-holonomic constraint as an equality pseudo-measurement~(GNSS\,+\, NHC), representing the state-of-the-art information-aided approach for land vehicle navigation~\cite{engelsman2023information}.
    \item \textbf{INQEKF}: To further motivate the dual-branch architecture, we evaluate only Branch 2, using GNSS updates with inequality constraints. That is, in this approach, the NHC equality constraints are not used.
\end{enumerate}
 
\subsection{Implementation Details}
Two primary scenarios are examined: full GNSS availability and GNSS-denied conditions. In both scenarios, the bounds for height, roll, and pitch were determined by taking the maximum absolute value of each variable according to the ground truth and multiplying it by a scale factor of two. These bounds were defined separately for each sequence. For the maximum forward velocity, a single fixed value of $50\,\mathrm{km/h}$ was used for all sequences, as detailed in \eqref{eq:d_v_max} and \eqref{eq:vd_max}.\\
Also, for all datasets, we added white noise with a standard deviation of 3.5 m to the GT positions in order to emulate a standard GNSS solution, and compared the results against the original RTK-GNSS measurements.\\
For the full GNSS availability scenario, we define a vector $\boldsymbol{\lambda} \in \mathbb{R}^{15}$ to compute the fusion weights, as described in \eqref{eq:w_propoused_inq}. The structure of $\boldsymbol{\lambda}$ is motivated by the covariance asymmetry between the two branches, as derived in Section~III-D. The specific numerical values for the $\boldsymbol{\lambda}$ vector were determined empirically on several randomly selected sequences and applied uniformly across all datasets.
\begin{equation}
\boldsymbol{\lambda} =
\begin{bmatrix}
0.85 & 0.85 & 10 & \mathbf{1}_{1 \times 2} & 10 & 10 & 10 & \mathbf{1}_{1 \times 7}
\end{bmatrix}^{\mathsf{T}}
\end{equation}
For the GNSS-denied scenarios, a one-minute initialization period with available GNSS measurements was performed to allow the filter to converge. After this initialization phase, the algorithm was executed for 30 seconds using only IMU data and the respective motion constraints.\\
For this scenario, a new weighting vector $\boldsymbol{\lambda} \in \mathbb{R}^{15}$ was defined to adjust the fusion weights accordingly. The horizontal position weights are reduced relative to the valid GNSS scenario, as the NHC equality constraint becomes the sole reliable aiding source during GNSS denial, and the fusion must prioritize its contribution in the unconstrained horizontal plane. Thus, we set
\begin{equation}
\boldsymbol{\lambda} =
\begin{bmatrix}
0.25 & 0.25 & 10 & 1_{\mathbf{1} \times 12}
\end{bmatrix}^\mathsf{T}
\end{equation}
\subsection{Full GNSS availability results}
\begin{figure}[t]
    \centering
    \includegraphics[width=\columnwidth]
    {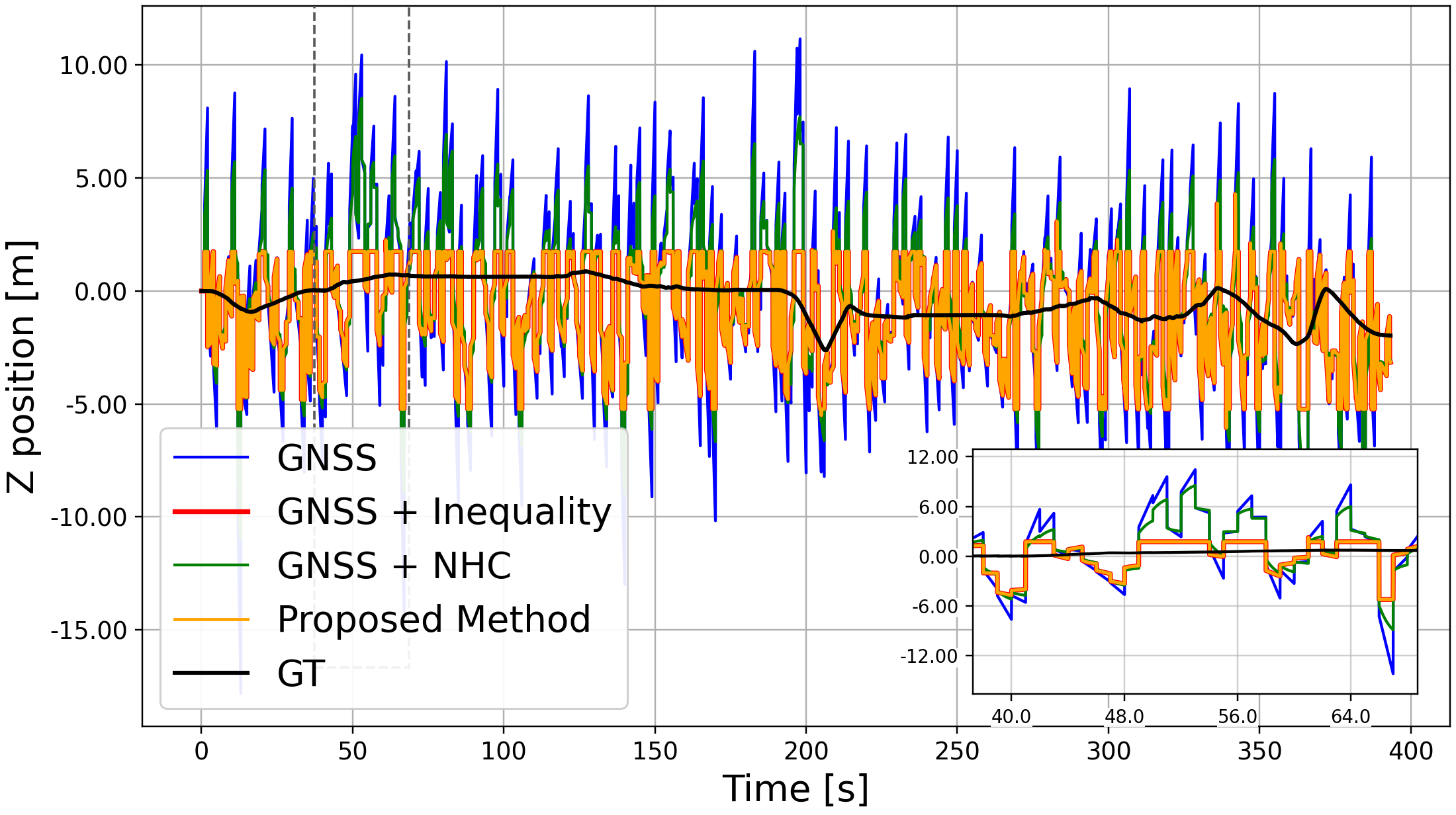}
    \caption{Vertical position over time for the Theater sequence 2 (Odyssey dataset).}
    \label{fig:z_error_valid}
\end{figure}

\begin{figure}[t]
    \centering
    \includegraphics[width=\columnwidth]
    {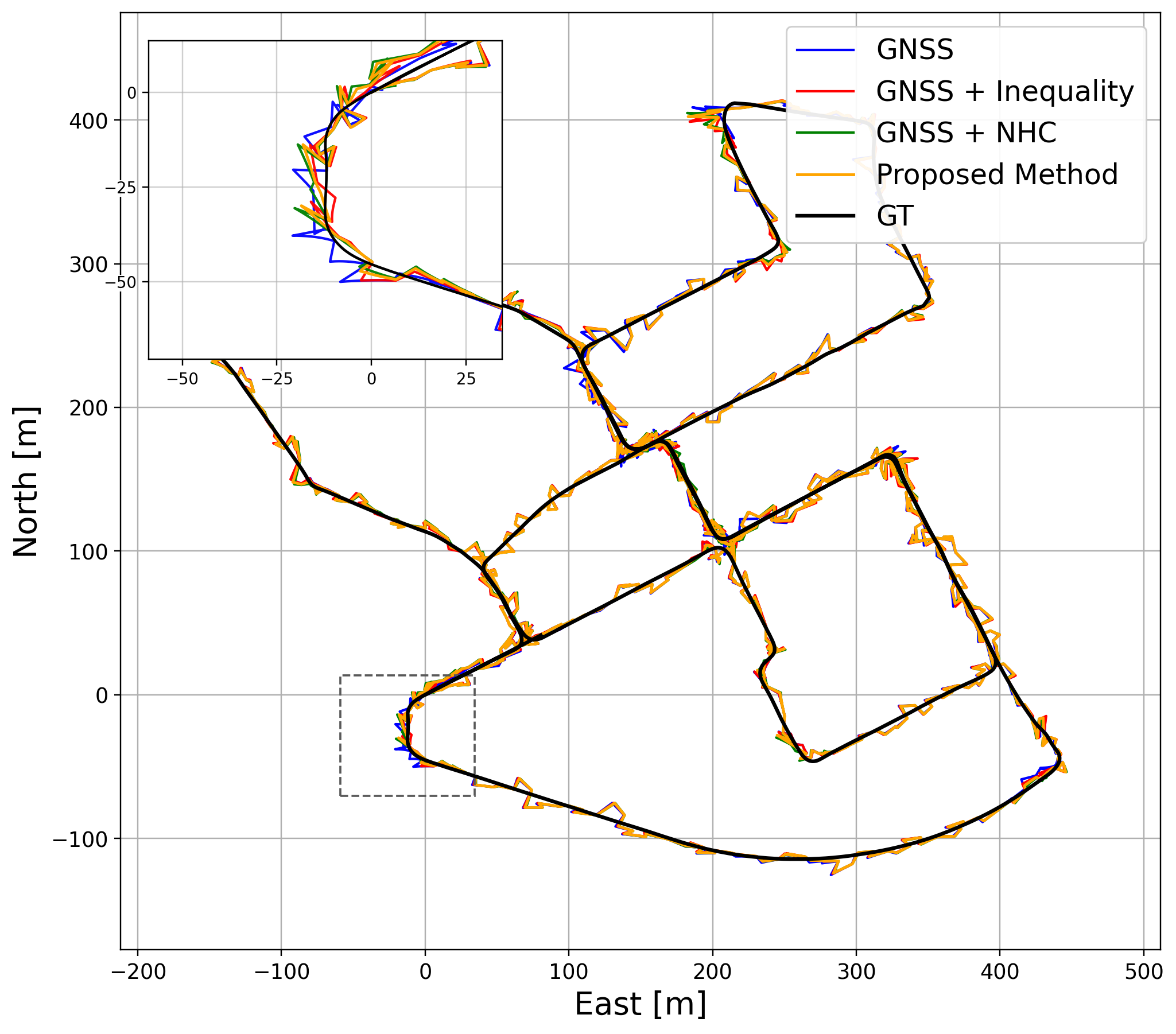}
    \caption{2D trajectory comparison for the residential sequence 
    27 at 10/03/2011 (KITTI dataset).}
    \label{fig:2d_map_valid}
\end{figure}

We evaluate all four approaches, EKF, NHCEKF, INQEKF, and the dual branch fusion, across all four datasets (Hong Kong, Odyssey, Tokyo, and KITTI) under full GNSS availability. For each dataset, we report the PRMSE \eqref{eq:prmse}, VRMSE \eqref{eq:vrmse}, ARMSE \eqref{eq:armse}, as well as the horizontal and vertical components of PRMSE separately. Results are averaged across all runs within each dataset, and an aggregate score over all datasets is reported to summarize overall performance.\\
The overall results are presented in Table~\ref{tab:rmse_valid_gnss}. The dual-branch fusion approach achieves consistent improvement across all evaluated datasets. The most significant performance gain across all evaluated metrics is observed in ARMSE, where the proposed method achieves an average ARMSE reduction of 50.1\% compared to NHCEKF. The NHCEKF enforces zero lateral and vertical body-frame velocity, which implicitly couples the attitude estimate to the velocity assumption. When this assumption is imperfectly satisfied, the equality constraint introduces attitude errors rather than correcting them. Indeed, in the Tokyo and Odyssey datasets, the NHCEKF itself degrades altitude accuracy relative to the EKF solution (14.82° vs 11.09° in Tokyo, 8.08° vs 4.32° in Odyssey), suggesting the NHC equality assumption is actively harmful for vehicle altitude in these sequences. We attribute this to their more dynamic driving profiles and, in the case of Tokyo, the lower IMU sampling rate (50 Hz), which reduces the accuracy of the body-frame velocity projection used in the NHC pseudo-measurement. The inequality constraints, by contrast, bound roll and pitch within their physically admissible range without imposing a rigid assumption on the nominal motion state, allowing the filter to correct attitude drift continuously even during dynamic maneuvers. The proposed method consequently improves the ARMSE over the EKF solution across all datasets as well, achieving an average ARMSE reduction of 31.8\%. This confirms that the attitude gains reflect a genuine improvement in attitude estimation, not merely recovery from a degraded NHCEKF.\\
To better understand the contribution to the position estimation, a detailed breakdown of horizontal and vertical position performance is provided in Table~\ref{tab:horizontal_vertical_position_valid_gnss}. The most pronounced gain is observed in vertical PRMSE, where an average PRMSE reduction of 16.7\% is obtained compared to NHCEKF. This result is expected, as the height bound directly constrains the vertical state, which is poorly observable from horizontal motion alone. In contrast, horizontal PRMSE improvement remains modest (1.0\%), confirming that the inequality constraints do not introduce artificial smoothing in the unconstrained plane.
\begin{table}[tb]
\centering
\footnotesize
\caption{RMSE under full GNSS availability. Percentages in parentheses indicate the improvement of our dual-branch approach over the other methods.}
\label{tab:rmse_valid_gnss}
\setlength{\tabcolsep}{3pt}
\renewcommand{\arraystretch}{1.0}
\begin{tabular}{lcccc}
\toprule
\textbf{Dataset} & \textbf{EKF} & \multicolumn{1}{c}{\textbf{NHCEKF}} & \multicolumn{1}{c}{\textbf{INQEKF}} & \textbf{Dual-Branch} \\
 & & & \textbf{(ours)} & \textbf{(ours)} \\
\midrule
\multicolumn{5}{l}{\textit{PRMSE [m]}} \\
\midrule
Hongkong & 6.80 & 6.48 & 6.22 & 6.17 \\
 & (9.3\%) & (4.8\%) & (0.9\%) & \\
Odyssey & 6.86 & 6.48 & 6.38 & 6.30 \\
 & (8.2\%) & (2.8\%) & (1.3\%) & \\
Tokyo & 7.04 & 6.99 & 6.65 & 6.54 \\
 & (7.1\%) & (6.5\%) & (1.7\%) & \\
Kitti & 6.95 & 6.73 & 6.40 & 6.36 \\
 & (8.5\%) & (5.4\%) & (0.6\%) & \\
\midrule
\textbf{All Datasets} & \textbf{6.91} & \textbf{6.67} & \textbf{6.41} & \textbf{6.34} \\
 & \textbf{(8.3\%)} & \textbf{(4.9\%)} & \textbf{(1.1\%)} & \\
\midrule
\multicolumn{5}{l}{\textit{VRMSE [m/s]}} \\
\midrule
Hongkong & 8.86 & 8.55 & 8.55 & 8.45 \\
 & (4.6\%) & (1.2\%) & (1.2\%) & \\
Odyssey & 4.51 & 4.36 & 4.04 & 3.96 \\
 & (12.3\%) & (9.2\%) & (2.1\%) & \\
Tokyo & 9.79 & 9.91 & 9.65 & 9.56 \\
 & (2.3\%) & (3.5\%) & (1.0\%) & \\
Kitti & 4.63 & 4.52 & 4.09 & 4.04 \\
 & (12.7\%) & (10.7\%) & (1.2\%) & \\
\midrule
\textbf{All Datasets} & \textbf{6.95} & \textbf{6.84} & \textbf{6.58} & \textbf{6.50} \\
 & \textbf{(6.4\%)} & \textbf{(4.9\%)} & \textbf{(1.2\%)} & \\
\midrule
\multicolumn{5}{l}{\textit{ARMSE [deg]}} \\
\midrule
Hongkong & 4.41 & 4.58 & 4.17 & 4.17 \\
 & (5.4\%) & (8.9\%) &  & \\
Odyssey & 4.32 & 8.08 & 4.32 & 4.31 \\
 & (0.4\%) & (46.6\%) & (0.1\%) & \\
Tokyo & 11.09 & 14.82 & 4.30 & 4.56 \\
 & (58.8\%) & (69.2\%) & (-6.1\%) & \\
Kitti & 5.16 & 6.65 & 4.05 & 3.99 \\
 & (22.8\%) & (40.0\%) & (1.5\%) & \\
\midrule
\textbf{All Datasets} & \textbf{6.25} & \textbf{8.53} & \textbf{4.21} & \textbf{4.26} \\
 & \textbf{(31.8\%)} & \textbf{(50.1\%)} & \textbf{(-1.1\%)} & \\
\bottomrule
\end{tabular}
\end{table}
\begin{table}[tb]
\centering
\footnotesize
\caption{Position RMSE under full GNSS availability. Percentages in parentheses indicate the improvement of our dual-branch approach over other methods.}
\label{tab:horizontal_vertical_position_valid_gnss}
\setlength{\tabcolsep}{3pt}
\renewcommand{\arraystretch}{1.0}
\begin{tabular}{lcccc}
\toprule
\textbf{Dataset} & \textbf{EKF} & \multicolumn{1}{c}{\textbf{NHCEKF}} & \multicolumn{1}{c}{\textbf{INQEKF}} & \textbf{Dual-Branch} \\
 & & & \textbf{(ours)} & \textbf{(ours)} \\
\midrule
\multicolumn{5}{l}{\textit{Horizontal Position RMSE [m]}} \\
\midrule
Hongkong & 5.61 & 5.52 & 5.59 & 5.53 \\
 & (1.3\%) & (-0.2\%) & (1.1\%) & \\
Odyssey & 5.71 & 5.73 & 5.71 & 5.61 \\
 & (1.6\%) & (2.1\%) & (1.7\%) & \\
Tokyo & 5.83 & 5.87 & 5.94 & 5.81 \\
 & (0.4\%) & (1.0\%) & (2.1\%) & \\
Kitti & 5.77 & 5.79 & 5.76 & 5.72 \\
 & (0.9\%) & (1.2\%) & (0.7\%) & \\
\midrule
\textbf{All Datasets} & \textbf{5.73} & \textbf{5.73} & \textbf{5.75} & \textbf{5.67} \\
 & \textbf{(1.1\%)} & \textbf{(1.0\%)} & \textbf{(1.4\%)} & \\
\midrule
\multicolumn{5}{l}{\textit{Vertical Position RMSE [m]}} \\
\midrule
Hongkong & 3.85 & 3.40 & 2.73 & 2.73 \\
 & (29.2\%) & (19.7\%) &  & \\
Odyssey & 3.82 & 3.03 & 2.86 & 2.86 \\
 & (25.1\%) & (5.5\%) &  & \\
Tokyo & 3.94 & 3.80 & 3.00 & 3.00 \\
 & (23.7\%) & (21.0\%) &  & \\
Kitti & 3.87 & 3.43 & 2.79 & 2.79 \\
 & (28.0\%) & (18.8\%) & & \\
\midrule
\textbf{All Datasets} & \textbf{3.87} & \textbf{3.42} & \textbf{2.84} & \textbf{2.84} \\
 & \textbf{(26.5\%)} & \textbf{(16.7\%)} &  & \\
\bottomrule
\end{tabular}
\end{table}
\\We also compare the 95th percentile error across the different approaches and observe significant improvements in vertical PRMSE and ARMSE, with aggregated gains of 18\% and 60\%, respectively. Importantly, these improvements do not degrade the performance of other state variables. Instead, we observe small but consistent improvements (approximately 1--2.5\%) in horizontal PRMSE and VRMSE when considering all datasets collectively. The detailed aggregated 95th percentile results are presented in Table~\ref{tab:percentile_95_valid_gnss}.\\
\begin{table}[tb]
\centering
\footnotesize
\caption{95th Percentile Error under full GNSS availability. Percentages in parentheses indicate the improvement ofour dual-branch approach over other methods.}
\label{tab:percentile_95_valid_gnss}
\setlength{\tabcolsep}{3pt}
\renewcommand{\arraystretch}{1.0}
\begin{tabular}{lcccc}
\toprule
\textbf{Metric} & \textbf{EKF} & \multicolumn{1}{c}{\textbf{NHCEKF}} & \multicolumn{1}{c}{\textbf{INQEKF}} & \textbf{Dual-Branch} \\
 & & & \textbf{(ours)} & \textbf{(ours)} \\
\midrule
Horizontal & 10.21 & 10.25 & 10.25 & 10.10 \\
 & (1.1\%) & (1.4\%) & (1.5\%) & \\
Position [m] & & & & \\
\midrule
Vertical & 7.66 & 6.74 & 5.52 & 5.52 \\
 & (28.0\%) & (18.1\%) &  & \\
Position [m] & & & & \\
\midrule
3D Velocity & 13.47 & 13.44 & 13.26 & 13.13 \\
 & (2.5\%) & (2.3\%) & (1.0\%) & \\
$\left[m/s\right]$ & & & & \\
\midrule
Altitude [deg] & 13.29 & 18.82 & 7.46 & 7.48 \\
 & (43.7\%) & (60.3\%) & (-0.3\%) & \\
\bottomrule
\end{tabular}
\end{table}
Fig.~\ref{fig:z_error_valid} illustrates the absolute vertical position error over time for run 2 of the Theater route (Odyssey dataset), demonstrating the consistent reduction in altitude error achieved by the proposed method relative to both baselines and highlighting the effectiveness of the height inequality constraint in suppressing vertical drift. In both figures, it can be observed that under full GNSS availability, the result of branch 2 (INQEKF) yields similar results to the dual-branch approach. Fig.~\ref{fig:2d_map_valid} presents the 2D trajectory for a single run of the residential sequence (KITTI dataset), where the proposed method (orange) closely follows the ground truth (black) throughout the route, illustrating the spatial consistency of the estimated path under full GNSS availability.\\

\subsection{GNSS-denied scenarios results}
\begin{figure}[t]
\centering
\includegraphics[width=\columnwidth]{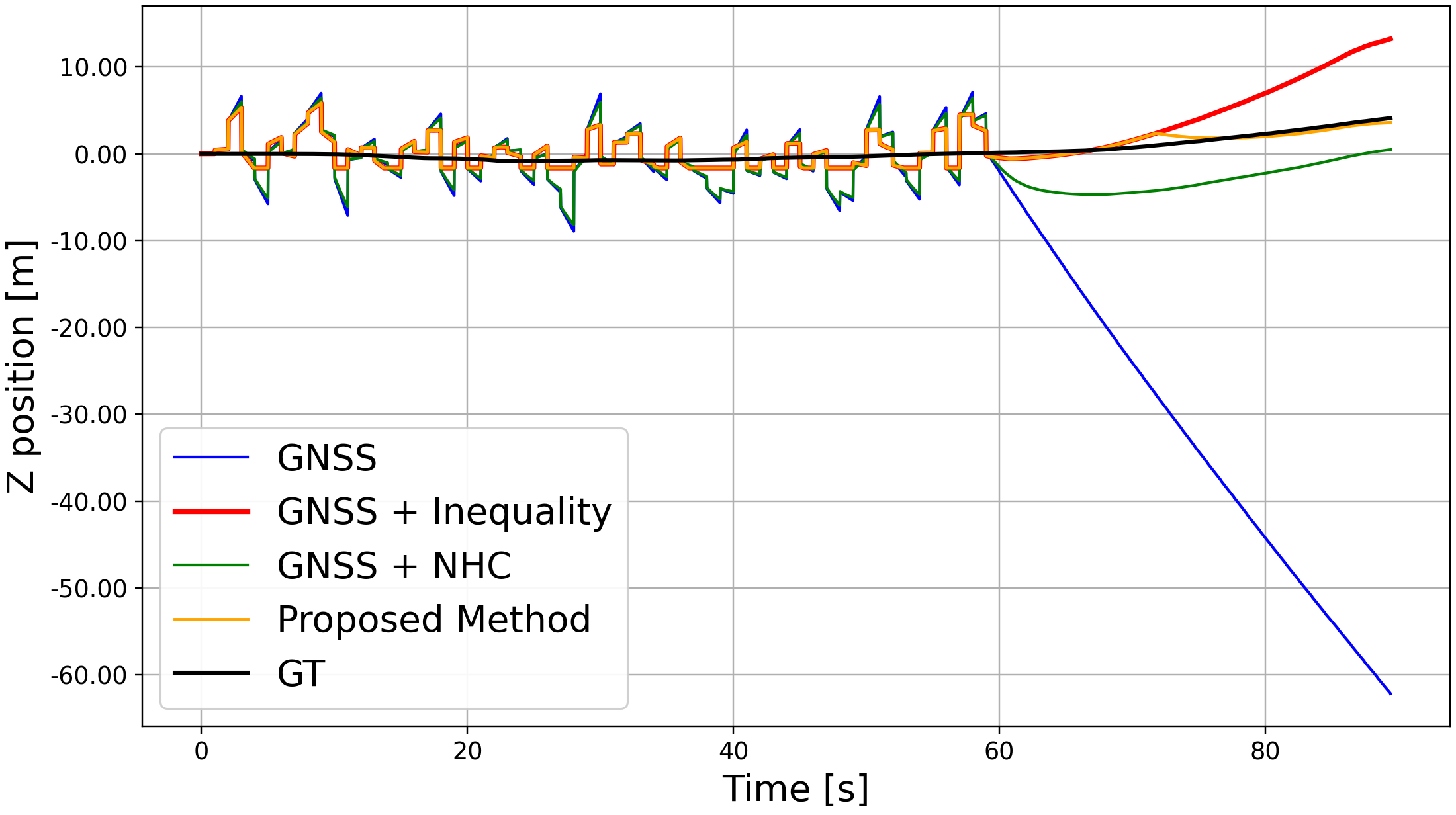}
\caption{Vertical position error during run of the Shinjuku route (Tokyo dataset).}
\label{fig:z_denied}
\end{figure}

\begin{figure}[t]
\centering
\includegraphics[width=\columnwidth]{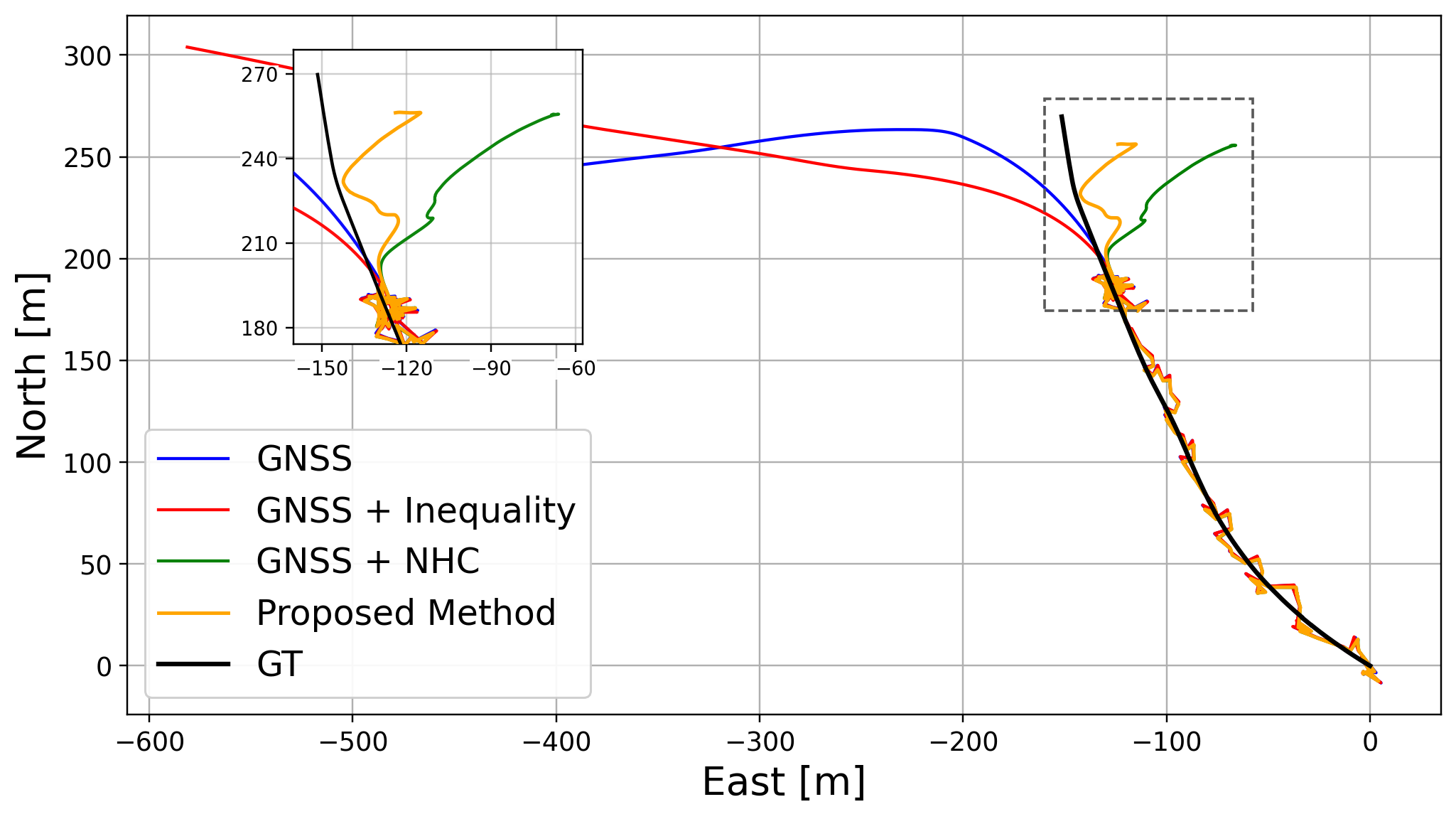}
\caption{2D trajectory during run 1 of the Theater route (Odyssey dataset) under GNSS outage conditions.}
\label{fig:2d_theater}
\end{figure}

We now evaluate the same four approaches under GNSS-denied conditions, simulating scenarios where satellite positioning is fully unavailable, and the system must rely solely on IMU integration aided by motion constraints. As expected, overall errors increase significantly compared to the full GNSS case.\\
The overall RMSE results are presented in Table~\ref{tab:rmse_gnss_denied}. The dual-branch fusion mitigates this degradation more effectively than alternative approaches. The average PRMSE is reduced from 113.64 [m] (EKF) to 60.39 [m], corresponding to a 3.6\% improvement over the next best method (NHCEKF). The VRMSE improves by an average of 7.0\%, with the most notable gains observed in the KITTI and Tokyo datasets, where reductions of 14.4\% and 13.5\% are obtained, respectively. These improvements reflect the ability of the inequality constraints to continuously bound IMU drift during dead-reckoning, even in the absence of any external aiding signal. ARMSE improves by an average of 20.2\% across all datasets.
An exception is observed in the Hong Kong dataset, where the proposed method yields a PRMSE of 33.01 [m] compared to 30.24 [m] for the NHCEKF solution, corresponding to a degradation of 9.1\%. Inspection of the dataset characteristics in Table~\ref{tab:dataset_parameters} reveals that the Hong Kong sequences exhibit the lowest height variation (8.73\,m average range) and the shortest total distance among all evaluated datasets. Under such flat, low-dynamic driving conditions, the height inequality constraint contributes minimally, while the variance-weighted fusion with the empirically determined 
$\boldsymbol{\lambda}$ vector biases the solution toward the inequality-constrained branch in a regime where the NHC equality assumption is well-satisfied. This behavior highlights a limitation of the proposed method: in environments where the equality constraints are consistently valid and vehicle dynamics are limited, the added complexity of the dual-branch fusion pipeline may offer no benefit over the NHCEKF solution.\\
A detailed breakdown of horizontal and vertical position components is provided in Table~\ref{tab:horizontal_vertical_position_gnss_denied}. Vertical position accuracy improves by an average of 24.2\% compared to the NHCEKF, with a particularly notable reduction of 37.8\% observed in the Tokyo dataset. Despite the anomaly observed in the Hong Kong dataset for PRMSE, the dual-branch fusion delivers strong and consistent gains in vertical accuracy across the remaining datasets, confirming that the height inequality constraint remains effective at suppressing altitude drift during dead-reckoning even without GNSS corrections.
\begin{table}[tb]
\centering
\footnotesize
\caption{RMSE under GNSS-denied scenarios. Percentages in parentheses indicate the improvement of our dual-branch approach over other methods.}
\label{tab:rmse_gnss_denied}
\setlength{\tabcolsep}{3pt}
\renewcommand{\arraystretch}{1.0}
\begin{tabular}{lcccc}
\toprule
\textbf{Dataset} & \textbf{EKF} & \multicolumn{1}{c}{\textbf{NHCEKF}} & \multicolumn{1}{c}{\textbf{INQEKF}} & \textbf{Dual-Branch} \\
 & & & \textbf{(ours)} & \textbf{(ours)} \\
\midrule
\multicolumn{5}{l}{\textit{PRMSE [m]}} \\
\midrule
Hongkong & 93.71 & 30.24 & 136.33 & 33.01 \\
 & (64.8\%) & (-9.1\%) & (75.8\%) & \\
Odyssey & 139.58 & 79.46 & 132.71 & 77.94 \\
 & (44.2\%) & (1.9\%) & (41.3\%) & \\
Tokyo & 96.93 & 46.11 & 103.40 & 40.06 \\
 & (58.7\%) & (13.1\%) & (61.3\%) & \\
Kitti & 124.35 & 94.85 & 99.49 & 90.54 \\
 & (27.2\%) & (4.5\%) & (9.0\%) & \\
\midrule
\textbf{All Datasets} & \textbf{113.64} & \textbf{62.67} & \textbf{117.98} & \textbf{60.39} \\
 & \textbf{(46.9\%)} & \textbf{(3.6\%)} & \textbf{(48.8\%)} & \\
\midrule
\multicolumn{5}{l}{\textit{VRMSE [m/s]}} \\
\midrule
Hongkong & 10.86 & 8.22 & 13.62 & 8.69 \\
 & (20.0\%) & (-5.8\%) & (36.1\%) & \\
Odyssey & 11.77 & 8.00 & 11.09 & 7.72 \\
 & (34.3\%) & (3.5\%) & (30.3\%) & \\
Tokyo & 9.73 & 9.81 & 9.71 & 8.49 \\
 & (12.8\%) & (13.5\%) & (12.6\%) & \\
Kitti & 11.53 & 9.38 & 9.77 & 8.03 \\
 & (30.4\%) & (14.4\%) & (17.9\%) & \\
\midrule
\textbf{All Datasets} & \textbf{10.97} & \textbf{8.85} & \textbf{11.05} & \textbf{8.23} \\
 & \textbf{(25.0\%)} & \textbf{(7.0\%)} & \textbf{(25.5\%)} & \\
\midrule
\multicolumn{5}{l}{\textit{ARMSE [deg]}} \\
\midrule
Hongkong & 5.46 & 5.29 & 5.50 & 5.11 \\
 & (6.4\%) & (3.4\%) & (7.1\%) & \\
Odyssey & 5.00 & 5.68 & 4.74 & 4.54 \\
 & (9.2\%) & (20.1\%) & (4.1\%) & \\
Tokyo & 2.50 & 2.28 & 2.56 & 2.09 \\
 & (16.2\%) & (8.4\%) & (18.4\%) & \\
Kitti & 5.18 & 6.68 & 4.26 & 4.16 \\
 & (19.6\%) & (37.6\%) & (2.3\%) & \\
\midrule
\textbf{All Datasets} & \textbf{4.53} & \textbf{4.98} & \textbf{4.27} & \textbf{3.98} \\
 & \textbf{(12.3\%)} & \textbf{(20.2\%)} & \textbf{(6.8\%)} & \\
\bottomrule
\end{tabular}
\end{table}
\begin{table}[tb]
\centering
\footnotesize
\caption{Position RMSE under GNSS-denied scenarios. Percentages in parentheses indicate the improvement of our dual-branch approach over other methods.}
\label{tab:horizontal_vertical_position_gnss_denied}
\setlength{\tabcolsep}{3pt}
\renewcommand{\arraystretch}{1.0}
\begin{tabular}{lcccc}
\toprule
\textbf{Dataset} & \textbf{EKF} & \multicolumn{1}{c}{\textbf{NHCEKF}} & \multicolumn{1}{c}{\textbf{INQEKF}} & \textbf{Dual-Branch} \\
 & & & \textbf{(ours)} & \textbf{(ours)} \\
\midrule
\multicolumn{5}{l}{\textit{Horizontal Position RMSE [m]}} \\
\midrule
Hongkong & 91.93 & 30.01 & 136.30 & 32.88 \\
 & (64.2\%) & (-9.6\%) & (75.9\%) & \\
Odyssey & 137.74 & 79.30 & 132.66 & 77.81 \\
 & (43.5\%) & (1.9\%) & (41.3\%) & \\
Tokyo & 93.33 & 45.89 & 103.36 & 39.96 \\
 & (57.2\%) & (12.9\%) & (61.3\%) & \\
Kitti & 122.58 & 94.62 & 99.36 & 90.41 \\
 & (26.2\%) & (4.4\%) & (9.0\%) & \\
\midrule
\textbf{All Datasets} & \textbf{111.39} & \textbf{62.46} & \textbf{117.92} & \textbf{60.27} \\
 & \textbf{(45.9\%)} & \textbf{(3.5\%)} & \textbf{(48.9\%)} & \\
\midrule
\multicolumn{5}{l}{\textit{Vertical Position RMSE [m]}} \\
\midrule
Hongkong & 18.16 & 3.73 & 2.67 & 2.94 \\
 & (83.8\%) & (21.3\%) & (-10.1\%) & \\
Odyssey & 22.64 & 5.06 & 3.77 & 4.50 \\
 & (80.1\%) & (11.1\%) & (-19.4\%) & \\
Tokyo & 26.20 & 4.41 & 2.92 & 2.74 \\
 & (89.5\%) & (37.8\%) & (6.2\%) & \\
Kitti & 20.89 & 6.58 & 5.12 & 4.82 \\
 & (76.9\%) & (26.7\%) & (5.8\%) & \\
\midrule
\textbf{All Datasets} & \textbf{21.97} & \textbf{4.94} & \textbf{3.62} & \textbf{3.75} \\
 & \textbf{(82.9\%)} & \textbf{(24.2\%)} & \textbf{(-3.6\%)} & \\
\bottomrule
\end{tabular}
\end{table}
Table~\ref{tab:percentile_95_gnss_denied} presents the 95th percentile error under GNSS-denied conditions. As expected, all methods exhibit substantial degradation compared to full GNSS availability conditions. However, the dual-branch fusion consistently reduces large-error occurrences relative to the baseline approaches.\\
In horizontal position, the dual-branch fusion achieves the lowest error (182.10\,m), corresponding to a 7.4\% improvement over the best competing method (NHCEKF). The most significant improvement is observed in vertical position, where the error is reduced to 6.22\,m, yielding a 42.5\% improvement relative to the next best approach. Additionally, the dual-branch fusion improves 3D velocity by 12.3\% and maintains competitive altitude accuracy, reflecting enhanced robustness against extreme deviations during GNSS outages.\\
\begin{table}[tb]
\centering
\footnotesize
\caption{95th Percentile Error under GNSS-denied scenarios. Percentages in parentheses indicate the improvement of our dual-branch approach over other methods.}
\label{tab:percentile_95_gnss_denied}
\setlength{\tabcolsep}{3pt}
\renewcommand{\arraystretch}{1.0}
\begin{tabular}{lcccc}
\toprule
\textbf{Metric} & \textbf{EKF} & \multicolumn{1}{c}{\textbf{NHCEKF}} & \multicolumn{1}{c}{\textbf{INQEKF}} & \textbf{Dual-Branch} \\
 & & & \textbf{(ours)} & \textbf{(ours)} \\
\midrule
Horizontal & 279.06 & 196.58 & 279.16 & 182.10 \\
Position [m] & (34.7\%) & (7.4\%) & (34.8\%) & \\
\midrule
Vertical & 61.96 & 10.83 & 7.58 & 6.22 \\
Position [m] & (90.0\%) & (42.5\%) & (17.9\%) & \\
\midrule
3D Velocity & 18.78 & 19.26 & 19.06 & 16.88 \\
$\left[m/s\right]$ & (10.1\%) & (12.3\%) & (11.4\%) & \\
\midrule
Altitude [deg] & 7.07 & 10.38 & 6.30 & 7.02 \\
 & (0.7\%) & (32.4\%) & (-11.4\%) & \\
\bottomrule
\end{tabular}
\end{table}
Fig.~\ref{fig:z_denied} illustrates the vertical position error during a single run of the Shinjuku route (Tokyo dataset), revealing the altitude divergence exhibited by all baselines during the outage. While all baselines diverge significantly, the dual-branch fusion maintains bounded vertical error throughout, confirming the effectiveness of the height inequality constraint as a standalone aiding source under GNSS-denied conditions. Fig.~\ref{fig:2d_theater} shows the 2D trajectory for run 1 of the Theater route (Odyssey dataset) during a GNSS outage, demonstrating the spatial drift behavior of each method in the absence of satellite corrections. Although all methods accumulate drift, the dual-branch fusion remains closest to the ground truth trajectory throughout the outage period, as highlighted in the inset. \\

\subsection{Summary}
The results reveal a clear and consistent pattern when comparing the two proposed methods (INQEKF and dual-branch) across the two evaluated scenarios (with or without GNSS aiding), and demonstrate consistent improvements over the state-of-the-art approaches (EKF and NHCEKF) under both nominal and GNSS-denied conditions.\\
Under full GNSS availability, the performance gap between INQEKF and the dual-branch fusion is marginal. In several cases, INQEKF achieves equal or slightly better results, most notably in vertical PRMSE, where both methods achieve identical scores across all datasets (2.84\,m average), and in ARMSE, where INQEKF outperforms the dual-branch fusion by an average of 0.05$^\circ$ (4.21$^\circ$ vs. 4.26$^\circ$). The dual-branch fusion achieves a synchronized improvement of 4.9\% in both PRMSE and VRMSE over NHCEKF, with PRMSE reduced from 6.91\,m to 6.34\,m and a peak improvement of 6.5\% in the Tokyo dataset. This suggests that when GNSS corrections are continuously available, the inequality constraints alone are sufficient to bound the state estimate, and the added complexity of the variance-weighted fusion with the NHC branch provides no significant benefit.\\
However, under GNSS-denied conditions, the distinction between the two methods becomes more pronounced, and the comparison reveals complementary strengths. The dual-branch fusion achieves the best overall PRMSE (60.39\,m) and VRMSE (8.23\,m), outperforming INQEKF (117.98\,m and 11.05\,m, respectively), which diverges significantly without GNSS corrections. This divergence is expected, as the inequality constraints alone cannot prevent horizontal drift in the absence of any absolute position reference, whereas the NHC equality constraint incorporated through Branch 1 provides a meaningful bound on lateral motion. In contrast, vertical position accuracy tells a different story: INQEKF achieves a lower average vertical PRMSE (3.62\,m) compared to the dual-branch fusion (3.75\,m), suggesting that the direct QP projection of the height constraint is more aggressive in suppressing altitude drift than the variance-weighted fusion, which partially dilutes the inequality branch contribution through the NHC branch weights.\\
A key characteristic of both proposed methods is their selective improvement of constrained dimensions. Horizontal position accuracy shows only marginal change (1.0\%), confirming that neither method introduces artificial smoothing. In contrast, ARMSE improves by 50.1\% overall (69.2\% in Tokyo), and vertical PRMSE improves by 16.7\% under full GNSS availability and by 24.2\% under GNSS denial. Notably, vertical PRMSE during GNSS outages is reduced from 21.97\,m to 3.75\,m, effectively suppressing altitude drift and significantly enhancing navigation robustness in challenging urban environments.\\
INQEKF is a strong standalone solution when GNSS is available, matching the dual-branch fusion in accuracy with lower architectural complexity. Under GNSS-denied conditions, neither method dominates across all dimensions: the dual-branch fusion is superior in horizontal position and velocity, while INQEKF provides tighter vertical bounding. The dual-branch fusion is therefore the preferred choice when overall navigation robustness across all state dimensions is required.

\section{Conclusion}
This paper presented a dual-branch EKF framework for INS/GNSS integration that incorporates both equality and inequality motion constraints through a variance-weighted fusion scheme. The results demonstrate that inequality constraints provide a complementary and robust source of aiding information that addresses a fundamental limitation of the standard NHC approach: its sensitivity to violations of the nominal motion assumption during dynamic maneuvers.\\
The key insight is that bounding the physically admissible range of velocity and attitude states, rather than enforcing a rigid equality assumption, allows the filter to benefit from kinematic constraints without being harmed when those constraints are only partially satisfied. This is reflected most clearly in the altitude results, where the dual-branch fusion consistently outperforms the NHCEKF across all datasets, including in scenarios where the equality constraint actively degrades accuracy relative to the unconstrained solution.\\
Under GNSS-denied conditions, the inequality constraints provide a meaningful check on vertical drift, which is the dominant error source in low-cost IMU dead-reckoning. The dual-branch fusion maintains bounded altitude error across all evaluated sequences, demonstrating practical utility for urban navigation in environments where GNSS outages are frequent and unpredictable.\\
The proposed approach is not without limitations. The $\boldsymbol{\lambda}$ vector, which governs the fusion weighting, requires empirical tuning of its specific numerical values, and the method offers reduced benefit in flat, low-dynamic environments where the NHC equality assumption is consistently satisfied, as demonstrated by the Hong Kong dataset results. Future work will investigate adaptive mechanisms for updating $\boldsymbol{\lambda}$ online based on real-time estimates of constraint reliability, and will explore extensions to non-flat terrain and higher-dynamic driving scenarios.\\
To conclude, we present a cost-free information aiding approach that improves navigation state estimation both with and without GNSS aiding. By exploiting physically admissible bounds on vehicle motion as soft inequality constraints, the proposed framework enhances robustness to dynamic maneuvers and mitigates vertical drift during GNSS outages, requiring only a software modification to an existing INS/GNSS integration filter.
\ifCLASSOPTIONcaptionsoff
  \newpage
\fi
\bibliographystyle{ieeetr}
\bibliography{references}

\end{document}